\def\year{2024}\relax
\documentclass[letterpaper]{article} 
\usepackage{aaai22}  
\usepackage{times}  
\usepackage{helvet}  
\usepackage{courier}  
\usepackage[hyphens]{url}  
\usepackage{graphicx} 
\usepackage{natbib}  
\usepackage{caption} 
\DeclareCaptionStyle{ruled}{labelfont=normalfont,labelsep=colon,strut=off} 
\usepackage{algorithm}
\usepackage{algorithmic}
\usepackage{booktabs}
\usepackage{tabularx}
\usepackage{multirow}

\usepackage{xcolor}
\newcommand{\answerYes}[1]{\textcolor{blue}{#1}} 
\newcommand{\answerNo}[1]{\textcolor{teal}{#1}}

\usepackage{newfloat}
\usepackage{listings}
\floatstyle{ruled}
\newfloat{listing}{tb}{lst}{}
\floatname{listing}{Listing}
\usepackage{tikz}
\usetikzlibrary{shapes.geometric, arrows}

\tikzstyle{process} = [rectangle, minimum width=5.5cm,  
minimum height=1cm, align=left, text width=5.4cm, draw=black, fill=orange!10]

\tikzstyle{arrow} = [thick,->,>=stealth]

\title{Keeping Humans in the Loop: \\ Human-Centered Automated Annotation with Generative AI}
\author{
    Nicholas Pangakis\textsuperscript{\rm 1},
    Samuel Wolken\textsuperscript{\rm 1}
}
\affiliations{
    \textsuperscript{\rm 1}University of Pennsylvania\\

}

\begin{document}

\maketitle

\begin{abstract}
Automated text annotation is a compelling use case for generative large language models (LLMs) in social media research. 
Recent work suggests that LLMs can achieve strong performance on annotation tasks; however, these studies evaluate LLMs on a small number of tasks and likely suffer from contamination due to a reliance on public benchmark datasets.
Here, we test a human-centered framework for responsibly evaluating artificial intelligence tools used in automated annotation. We use GPT-4 to replicate 27 annotation tasks across 11 \textit{password-protected} datasets from recently published computational social science articles in high-impact journals. For each task, we compare GPT-4 annotations against human-annotated ground-truth labels and against annotations from separate supervised classification models fine-tuned on human-generated labels. Although the quality of LLM labels is generally high, we find significant variation in LLM performance across tasks, even within datasets.  Our findings underscore the importance of a human-centered workflow and careful evaluation standards: Automated annotations significantly diverge from human judgment in numerous scenarios, despite various optimization strategies such as prompt tuning. Grounding automated annotation in validation labels generated by humans is essential for responsible evaluation.

\end{abstract}

\noindent

\section{Introduction}

\begin{figure}
    \centering
    \resizebox{\columnwidth}{!}{
    \begin{tikzpicture}

    \node[inner sep=0pt] (human) at (-3,0)
        {\includegraphics[width=.15\textwidth]{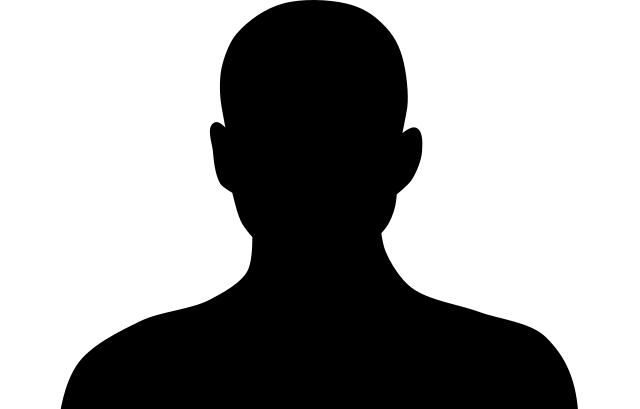}};
    \node[below, align=center, font=\small] at (-3, 1.3) {Human Annotator};
    
    \node[inner sep=0pt] (robot) at (1.4,2.5)
        {\includegraphics[width=.1\textwidth]{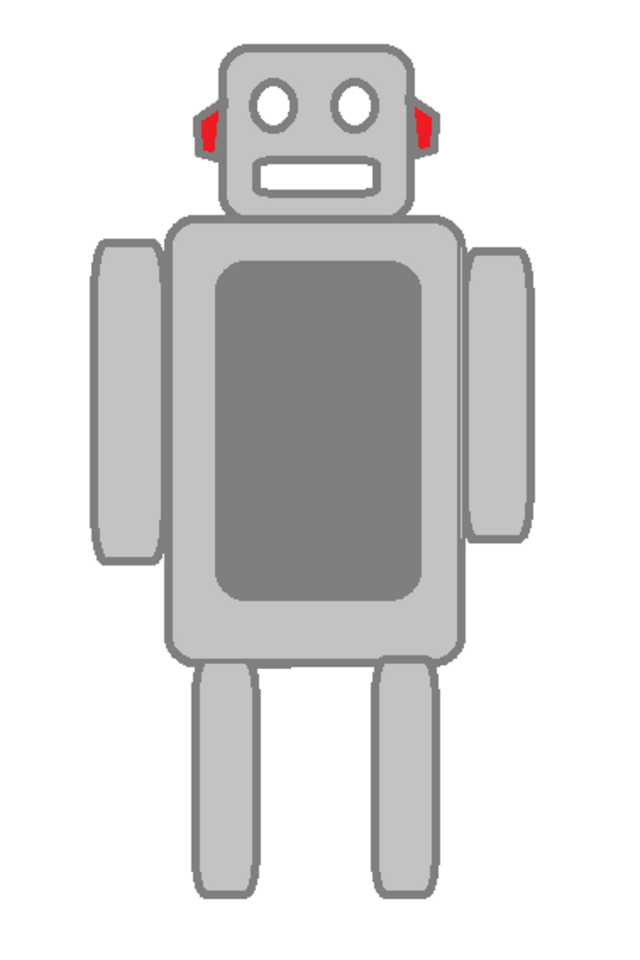}};
        \node[below, align=center, font=\small] at (2.9, 3) {Generative \\ LLM};

    \node[draw, draw=orange!10, minimum width=2cm, minimum height=1cm,fill=orange!10,font=\small,align=center,rounded corners] (box1) at (-3, 2.49) {1) Create task-specific instructions\\to serve as the LLM prompt\\ and human codebook};

    \node[draw, draw=blue!10, minimum width=2cm, minimum height=1cm,fill=blue!10,font=\small,align=center,rounded corners] (box2) at (0, 0) {2) Validate LLM\\on human-\\labeled subset};
    
    \node[draw,draw=blue!10, minimum width=2cm, minimum height=1cm, fill=blue!10,font=\small,align=center, rounded corners] (box3) at (3, 0) {3) Prompt\\optimization};

    \node[draw, draw=green!10,minimum width=2cm, minimum height=1cm,fill=green!10,align=center,font=\small] (box4) at (1, -2) {4) Using optimized prompt, test LLM performance \\  on remaining human-labeled samples};
    

    \draw[->, ultra thick] (-2.25, 0) -- (box2.west);
    \draw[->,thick] (box1.south) -- (-3,1.3);
    \draw[->,thick] (box1.east) -- (0.63,2.5);
        
    \draw[->,thick]  (1.2,-.2) -- (1.98,-.2);
    \draw[->,thick]  (1.98,.2) -- (1.2,.2);
    
    \draw[->, dashed, ultra thick] (.75, 1.5) to[out=190, in=90] (box2.north);

    \draw[->, dashed, ultra thick] (2, 1.5) to[out=0, in=90] (box3.north);

    \draw[->, ultra thick] (human.south) to[out=270, in=180] (box4.west);
    \draw[->,dashed,  ultra thick] (box3.south) -- (3, -1.5);
\end{tikzpicture}}
    \caption{Human-in-the-loop workflow for augmenting text annotation with generative LLMs}
    \label{fig:enter-label}
\end{figure}
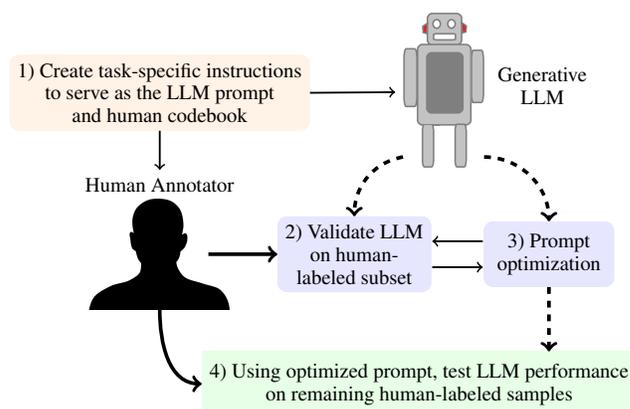

Scholars studying social media routinely employ natural language processing (NLP) tools to analyze large quantities of text data. Classification is among the most common NLP tasks, allowing researchers to describe the text of social media posts at scale. Social media researchers have begun to deploy generative LLMs for annotation for important tasks, such as identifying hate speech \citep{kumarage2024harnessing}, public attitudes towards vaccines \citep{espinosa2024use}, and measuring news outlet credibility \citep{yang2023large}. Classification tasks depend on high-quality, manually-labeled text data for training and validation of supervised and semi-supervised language models \citep{grimmer22}. Computational social scientists typically use the labeled text data from these models for descriptive purposes or to serve as control, predictor, or outcome variables in statistical analyses \citep{egami22, knox22, roberts}.

Here, we explore the potential of generative large language models (LLMs) to automate the manual annotation process as a few-shot classification problem \citep{NEURIPS2020_1457c0d6}. LLMs are faster and cheaper than human annotators and do not suffer from human shortcomings such as limited attention span or fatigue \citep{grimmer2013, Neuendorf16}. As a result, LLMs may be valuable tools for annotation.
\looseness=-1

In Figure 1, we propose a human-centered workflow for automated annotation. Despite inevitable imperfections of human annotators, we argue that deployment of artificial intelligence (AI) technology for automated annotation \textit{must} remain grounded in human judgment. Human-centered artificial intelligence (HCAI) is a central principle of responsible AI and a critical component to developing automated technology that is Reliable, Safe, and Trustworthy (RST) \citep{falco, hcai}. Without a human-centered approach that relies on human validation data, researchers cannot identify and measure biases in annotation performance.
\looseness=-1

A growing body of research on automated annotation claims that generative LLMs can match or exceed human performance on annotation tasks \citep{chiang2023can, ding2022gpt3, he2023annollm, mellon22, pan, thapa, tornberg, tornberg2, zhu2023chatgpt}. In tension with a human-centered approach, some of this research is framed to evaluate whether LLMs are ``better" than human annotators \citep[e.g.][]{better}. For instance, \citet{gilardi2023chatgpt} conclude that LLMs \textit{outperform} typical human annotators: ``[t]he evidence is consistent across different types of texts and time periods. It strongly suggests that ChatGPT may already be a superior approach compared to crowd annotations on platforms such as MTurk.'' Others argue that automated annotations by LLMs  may be higher quality than annotations by human experts \citep{tornberg} and, as a result, ``human coding hence cannot be treated as an unquestioned gold standard" \citep{tornberg2}. The implication is that generative LLMs could \textit{replace} human annotators.

Despite these claims, it is unclear whether LLM performance in previous studies can generalize to other datasets and tasks. Most studies in this vein analyze a few insular tasks on a small number of datasets test LLM performance on publicly available benchmark datasets \citep{wang, ziems23}, which are plausibly affected by data leakage and contamination.\footnote{Contamination means that the datasets may be included in the LLM's pretraining data.} As a result, strong performance may reflect memorization, which will not generalize to datasets and tasks that are not included in the LLM's training data.  Aside from concerns about contamination, errors in automated annotation are likely to be correlated across samples, which may lead to systematic biases in labels \citep{ashwin2023using}. Others have noted that inconsistency in ChatGPT's annotations means they are unreliable \citep{Ollion, reiss2023testing}. These concerns suggest that rigorous, task-by-task validation is essential for practitioners relying on LLMs for automated annotation. 

If the strong performance demonstrated in recent work generalizes to novel text annotation tasks in social science, then validation may not be a significant concern. If performance \textit{does not} generalize to a wide range of tasks, then researchers risk generating inaccurate labels that will introduce bias in downstream description or analysis. Researchers who use LLMs for automated annotation without human-centered validation assume LLM annotations align with human judgment about the conceptual meaning of categories. Without ground-truth labels, these assumptions cannot be tested. We argue that using LLMs for automated annotation responsibly and effectively requires a human-centered approach.

In this paper, we highlight the importance of grounding automated annotation in human judgment by testing the generalized performance of automated annotation on tasks and datasets less likely to be affected by contamination and leakage. We use generative LLMs to replicate 27 different manual annotation procedures from 11 non-public datasets from computational social science (CSS) articles recently featured in high-impact publications. The replicated tasks represent a range of real-world applications in CSS. Each of the original datasets includes human-labeled annotations that we treat as ground-truth.\footnote{We consider the human labels from the original studies as ground truth because the annotation procedures underwent independent review and scrutiny during the peer-review process and were determined to be quality enough for publication in a high-impact journal. All of the annotation procedures in the original studies employed either multiple annotators or an expert coder. We discuss the human annotation procedures from the original studies in Appendix B.} Rather than testing whether LLMs ``outperform" human annotators, we explore how well they can approximate human judgment. We ask the following question: Can generative AI reasonably approximate human annotations in computational social science?

For each task, we provide GPT-4\footnote{We selected GPT-4 because it was the highest performing generative LLM at the time of our analyses.} with detailed instructions to label text samples into conceptual categories outlined in the original study. We then conduct direct label-to-label comparisons between the GPT-4 annotations and the human annotations in the original study. To compare these results to a baseline, we also fine-tune several supervised text classifiers for every task. Unlike the GPT-4 few-shot approach, these supervised models were fine-tuned with human-generated labels only. Due to its low cost, ease of application, and high performance, we fine-tune BERT classifiers \citep{bert} on varying training sample sizes and use them to benchmark GPT-4's performance. After our primary investigation, we conduct a series of secondary analyses aimed at exploring ways to improve automated annotation performance. These experiments include testing prompt optimization, temperature tuning, and a novel strategy to identify lower confidence automated annotations. 

We identify four main contributions from our results:
\begin{enumerate}
\item GPT-4 text annotation performance is promising but inconsistent across tasks and datasets. Despite excelling on many tasks, automated annotation significantly diverges from human judgment in numerous scenarios. Across all replicated tasks, we find a median accuracy of 0.850 and a median F1 of 0.707. Nevertheless, nine of the 27 tasks were below 0.5 on precision or recall. For a full one third of the sampled tasks, automated annotation would result in labels for which false positions outnumber true positives or more than half the true positives are missed. Grounding automated annotation in validation labels generated by humans is essential for responsible evaluation.
\item GPT-4 performance is significantly stronger in \textit{recall} than \textit{precision}. For 20 of the 27 tasks, recall exceeds precision. Thus, one valuable use case of automated annotation may be as a first stage in a multi-stage annotation pipeline, where human annotators manually review all positive class instances from the LLM outputs. 
\item Various strategies to improve automated annotation performance, including manual prompt optimization and tuning LLM hyperparameters like temperature, improve annotation performance only marginally. These strategies are unlikely to solve the inconsistencies in poor LLM annotation performance at this time.
\item With adequate training data, supervised encoder-only LLM classifiers surpass GPT-4 performance. GPT-4 only outperforms supervised classifiers when there are minimal training samples.

\end{enumerate}

\section{Data and Replication Procedures}

\subsection{Data}

The data used in our replication analyses include 27 manual annotation tasks from 11 CSS articles recently published in high-impact journals.\footnote{Table A1 lists the complete list of the datasets and Table A2 details each annotation task.} These tasks were selected to represent a range of annotation tasks in contemporary CSS research, drawing from research appearing in interdisciplinary publications (e.g., \textit{Science Advances} and \textit{Proceedings of the National Academy of Sciences}) as well as high-impact field journals in political science (e.g., \textit{American Political Science Review} and \textit{American Journal of Political Science}) and psychology (e.g., \textit{Journal of Personality and Social Psychology}). We identified these articles by searching journals for articles that utilized manual annotation procedures in their research design. All replicated annotation procedures appeared in articles that were published within the last three years and included replication data in a password-protected data repository. 

We treat each label in the original datasets as a separate binary classification task. While the majority of the classification tasks we replicate were originally binary classification tasks, we decompose the remaining multiclass classification tasks into binary tasks. Decomposing classification problems into a series of binary classification problems is a simple, common approach for multiclass classification \citep{allwein2000reducing,joshi2017aggressive,lorena2008review}. This approach has several advantages for our analysis. First, it harmonizes how we execute and evaluate each replicated task, allowing for a standardized classification workflow across tasks. Second, treating each category as a separate binary classification task allows for more granular error analysis for LLM performance. Given our focus on evaluating LLMs at text annotation tasks, decomposing single complex tasks into several simpler tasks gives more opportunities for insight into heterogeneity in LLM annotation performance.

Our tasks cover varying degrees of class imbalance: Across all tasks, the minimum positive class frequency is 0.04\%, the median is 24.5\%, and the maximum is 70.7\%. The sources of the human labels are representative of common approaches to human annotation: 22.2\% of tasks were annotated by crowdsourced workers, 37\% by experts, and 40.8\% by research assistants. The annotation tasks reflect an extensive range of CSS applications, from identifying whether Cold War-era texts pertained to political affairs or military matters \citep{schub22} to analyzing open-ended survey responses to classify how people conceptualize where beliefs come from \citep{Cusimano20}. 

Because we rely exclusively on datasets stored in password-protected data archives (e.g., Dataverse) or datasets secured through direct outreach to authors, our replication analyses are unlikely to be affected by contamination and data leakage. Task and test data contamination are significant, yet understudied, challenges inherent to evaluating LLMs \citep{NEURIPS2020_1457c0d6, li, magar}. Because LLMs are trained on huge volumes of text data, using digital resources such as the Common Crawl, they are able to ``memorize'' publicly available information during the training process. This issue is compounded due to the fact that the highest performing models (e.g., GPT-4) are closed-source LLMs and the proprietors do not release details on pre-training data. Although open-source models provide more detail on their training data, extracting data from crawled websites can be difficult and prone to drift over time \citep{li}. By analyzing public data and tasks released before and after the LLM pre-training period, \citet{li} determine that contamination is a widespread evaluation issue and conclude that,``[d]ue to task contamination, closed-sourced models may demonstrate inflated performance in zero-shot or few-shot evaluation, and are therefore not trustworthy baselines in these settings.'' 

Prior research that examines automated annotation on a range of tasks mainly evaluates their models on popular, publicly available benchmarks released before the LLM training process \citep{wang,ziems23}. Therefore, it is not clear whether these findings will generalize to non-contaminated tasks, which are more representative of CSS tasks.

\subsection{Replication Procedures}

We follow the four-step human-in-the-loop workflow described in Figure 1 to replicate each annotation task. This workflow centers human judgment by using human ground-truth labels to guide prompt optimization and to evaluate label quality.

In step 1, we create instructions (i.e., a qualitative codebook) for each task that delineates the conceptual categories of interest.\footnote{Because LLMs respond to natural language prompts, a clear set of instructions is essential for LLM annotation tasks. A rich literature spanning both qualitative and quantitative social science offers guidance on how to develop a codebook and classify text data based on relevant concepts \citep[e.g.,][]{krip}.} 
We develop the LLM prompts from the exact human annotator instructions used in the original research design. If the instructions were not directly available, we either quote or paraphrase text from the article or supplementary materials that describes the concepts of interest.\footnote{We do not observe any relationship between LLM performance and whether or not the verbatim wording used to generate labels  was available. In the supplementary material, we provide all prompt instructions.} 

For step 2, we have the LLM label a randomly selected subset (n=250) of text samples already labeled by humans using the human codebook as a prompt.\footnote{If humans and LLMs label the data using different codebooks, there could be a conceptual gap between the two annotation instructions, which could lead to a threat in content validity \citep{jacobs2021measurement}.} To ground automated annotation in human judgment, we then calculate classification performance metrics (i.e., accuracy, recall, precision, and F1) by comparing the generative LLM predicted labels against the ground-truth human labels.

In step 3, we refine the prompts by reviewing patterns in errors made by the LLM and by incorporating examples from incorrect LLM classifications.\footnote{One concern at this stage is overfitting the prompt instructions to this particular subset of data. Moreover, if fundamental changes were made to the prompt instructions, it may be prudent to have the humans re-label the original text samples as well.} To do so, we use the same 250 samples from step 2. This step is intended to optimize the quality of the prompt instructions by making LLM judgment align as closely as possible to human judgment. 

In step 4, we use the generative LLM to label a separate set of 1,000 held-out text samples with the updated prompts. We assess final performance against these held-out samples. Performance on this held-out set can be used to determine whether an LLM can be reasonably utilized for automated annotation. The code made available in our GitHub repository offers a simple and efficient way to implement the procedures outlined in the above workflow.\footnote{See: \url{https://github.com/npangakis/gpt_annotate}} 

{
\begin{table}
    \centering
    \begin{tabular}{lccccc}
    \toprule
    Metric & Minimum  & Mean & Median   & Maximum \\
    \midrule
    F1 & 0.06  & 0.66 &  0.71  & 0.97 \\
    Accuracy &  0.67  & 0.86 &  0.85  & 0.98 \\
    Precision & 0.03  & 0.62 &  0.65  & 0.96 \\
    Recall & 0.25  & 0.75 & 0.83 & 0.98 \\
    \bottomrule
    \end{tabular}
    \caption{Automated annotation performance across 27 tasks from 11 datasets, assessed on 1000 human-generated labels.}
    \label{tab:metrics}
\end{table}
}

Across all 27 tasks, we annotate slightly over 75,000 text samples using OpenAI's \textsc{GPT-4} API.\footnote{Scholars have raised concerns about relying on LLMs in social science research due to the constant evolution of LLM software, the black-box nature of how LLMs process queries, and ambiguity about the training data that underlies LLMs. Open-source LLMs represent a viable solution to many of these problems \citep{spirling2023}. The workflow outlined here is LLM-agnostic and could be adapted to any open-source LLM.} The overall cost was under \$500 USD. On average, a dataset with 1,000 text samples took approximately 1 hour to label using GPT-4. Together, the low cost and relatively rapid speed demonstrate the potential value of LLM-augmented annotation in CSS.  

To set baselines for model performance, we also replicate each of the annotation tasks as a supervised text classification problem using an encoder-only LLM. For our analyses, we classify the same held-out set from step 4 above using BERT models fine-tuned on varying sample sizes of human-labeled text samples. To tune each model's hyperparameters, we train 18 separate classifiers with varying hyperparameters and select the model with the highest F1 performance.\footnote{We optimize the learning rate, the batch size, and the number of epochs. We elaborate on this process in Appendix C.} In total, we trained over 500 supervised LLM classifiers to better understand the performance of GPT-4 automated annotation relative to alternative classification strategies.

Finally, after conducting our main analyses, we test a variety of approaches to improve automated annotation performance, including prompt optimization, temperature tuning, and a novel selection strategy. We also explore whether GPT-4 performance is change over time. These secondary experiments aim to understand whether there may be specific ways to overcome poor performing tasks.

\section{Results}
Automated annotation performance metrics are shown in Table \ref{tab:metrics}. The performance metrics reported are based on 1,000 ``held out'' text samples per task (i.e., text samples not used in the prompt optimization process from step 2 of the workflow). Across the 27 tasks, GPT-4 achieved a median F1 score of 0.707 and median accuracy of 0.85.\footnote{Prior work (e.g., Gilardi, Alizadeh, and Kubli 2023) also reports LLM “intercoder agreement,” which indicates how consistently an LLM labels a text sample. Yet, this measure cannot distinguish between inconsistency due to poor performance, substantive ambiguity, or arbitrary randomness governed by a “temperature” parameter. As a result, we do not leverage this metric for automated annotation evaluation. Instead, we exploit this measure as a gauge of annotation ``confidence," which we explore in our “Exploiting Uncertainty in LLM Labels” section.} Thus, automated annotation is a viable strategy for a significant portion of the replicated tasks.\footnote{Appendix D investigates an open-source alternative to the methods explored here.}   

\begin{figure}[H]
    \centering
    \includegraphics[width=.95\linewidth]{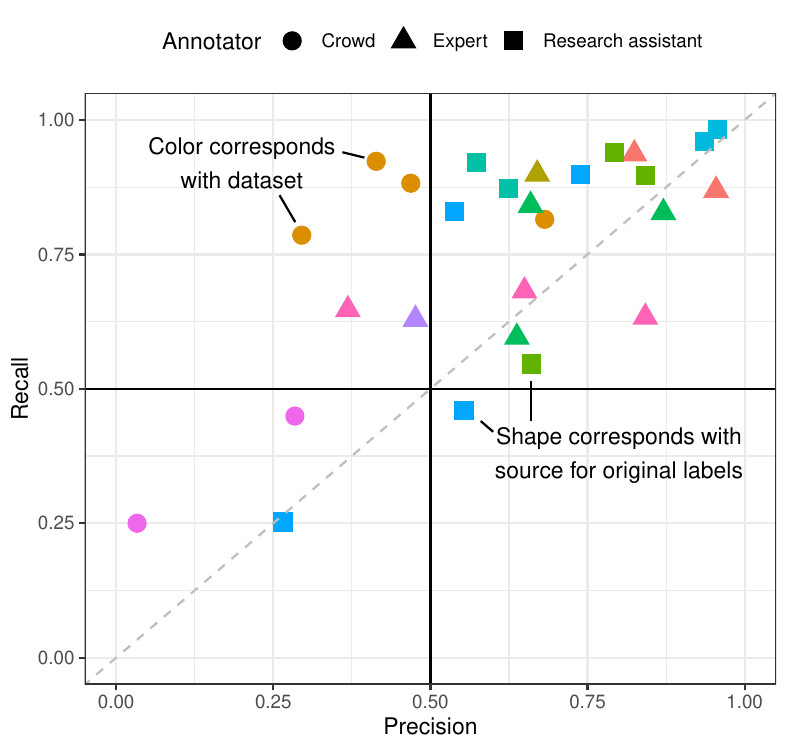}
    \caption{Precision and recall for 27 annotation tasks compared to human labels. Points sharing the same color are conducted on the same text data set.}
    \label{fig:quadrant}
\end{figure}

Figure \ref{fig:quadrant} shows precision and recall for each annotation task. On eight of the 27 tasks, the LLM achieves remarkably strong performance with precision and recall both exceeding 0.7. Nevertheless, automated annotations diverged from human judgment in a non-trivial number of cases. Specifically, nine of the 27 tasks had either precision or recall below 0.5---and three tasks had both precision and recall below 0.5. Put otherwise, for a full one-third of tasks, the LLM either missed at least half of the true positive cases, had more false positives than true positives, or both. At worst, the aggregate performance ranged as low as an F1 score of 0.06. These results are not driven by one data set or type of annotation source. Moreover, LLM performance varied substantially across tasks within a single dataset. In the most extreme example, F1 ranged from 0.259 to 0.811 on two separate tasks a single dataset \cite[political speech data from][]{card23}, a difference of 0.552. These results demonstrate the variability of automated annotation performance and, accordingly, underscore the need for task-specific human validation.

\begin{figure*}
    \centering
    \includegraphics[width = .8\textwidth]{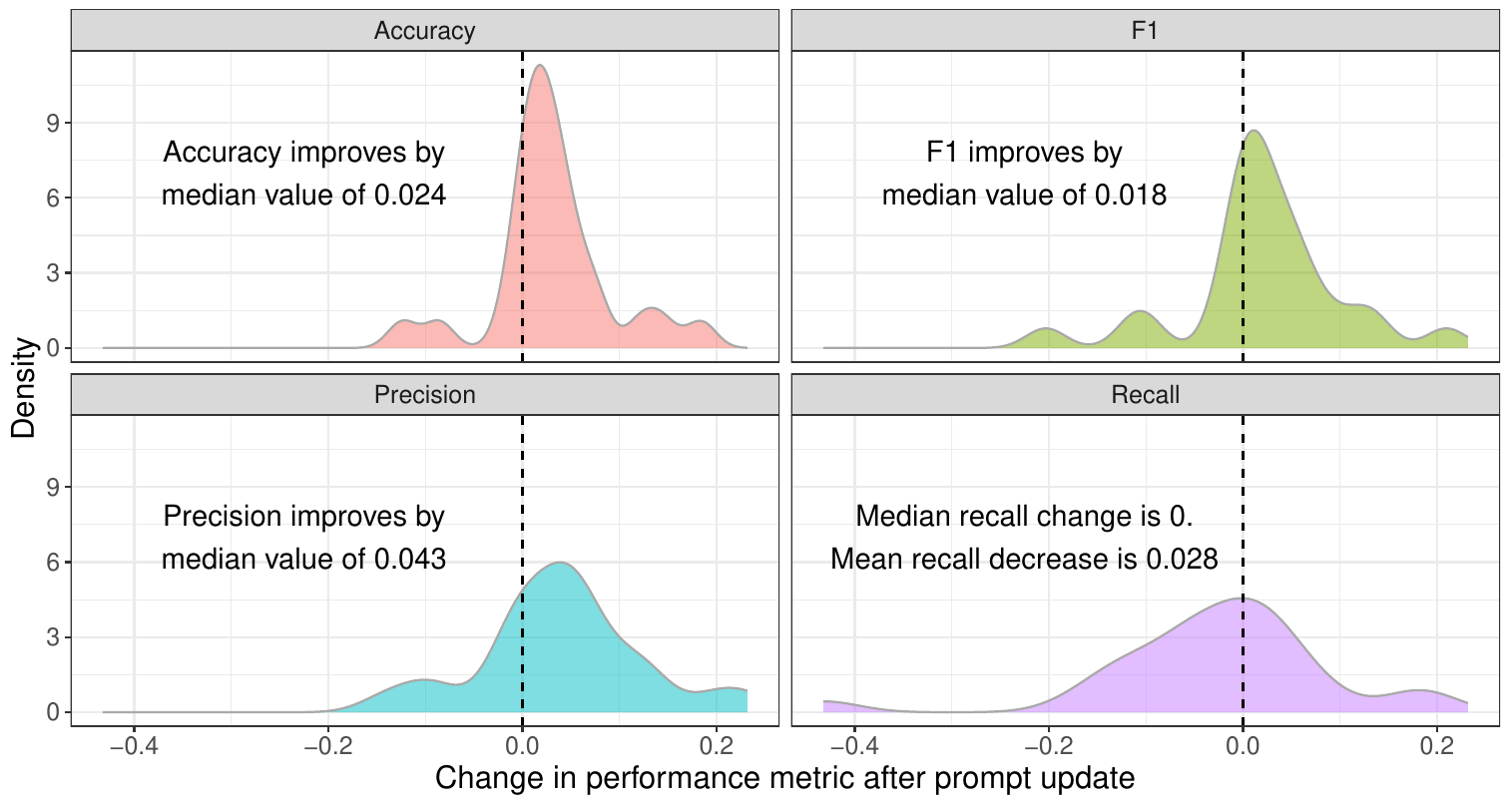}
    \caption{Change in LLM annotation performance on training data after one round of prompt updates.} 
    \label{fig:codebook_updates}
\end{figure*}

Although performance was strong across many tasks, especially those akin to summarization \citep[for instance][]{muller}, it suffered on more abstract tasks. For instance, poor performance correlated with tasks relying on nuanced cultural knowledge, such as differentiating between national, racial, and religious identities \citep{hopkins}. Subpar performance also occurred when the task required inferences about conversational context \citep{peng}. Although we cannot be certain why certain tasks performed poorly, it is possible that, in general, performance was lower than prior studies because the data sets analyzed here were stored in password-protected archives and thus not affected by memorization. In Appendix E, we conduct additional analyses investigating how performance varies across different conceptualizations of task and text difficulty.

As is clear from Figure \ref{fig:quadrant}, LLM performance is stronger on recall than precision for 20 of the 27 tasks. Overall, the median recall (0.83) is significantly higher than the median precision (0.65). Accordingly, automated annotation may be especially valuable as a first stage in multi-stage annotation pipelines, providing high-recall preliminary labels that are refined in subsequent stages by models optimized for precision or by additional manual review. Following a human-machine collaborative framework \citep{coannotate}, one future application may be to use generative AI as a technique to create more balanced classes before the manual annotation process. This would involve an LLM labeling a large number of text samples and then downsampling to create class balance for the initial human annotation stage.

\subsection{Manual Prompt Optimization}

For each of the 27 annotation tasks, we made iterative human-in-the-loop updates to the instructions to optimize the prompt for accurate annotations. Figure 3 shows the distributions of change in performance metrics after updating the prompt and re-annotating the same text samples. This analysis tests whether and how the prompt update process affects LLM annotation, holding constant the data and conceptual categories. 

In most cases, the prompt update process led to modest increases in accuracy and F1---although improvements were largely driven by changes in precision. While the magnitude of improvement was generally small, researchers experiencing subpar LLM annotation performance can use human-in-the-loop prompt refinement to ensure that their instructions are not the cause of poor performance. Because the improvements are small, however, it is unlikely that prompt tuning will completely resolve poor automated annotation performance at this time. Either way, a human-centered framework will remain essential: Even with improved LLM technology and automated prompt optimization strategies \citep{khattab2023dspycompilingdeclarativelanguage}, researchers will always require human-labeled ground truth text data to optimize the instructions.

\subsection{Exploiting Uncertainty in LLM Labels}

LLMs may be an effective way to identify text samples that cannot be clearly classified into the annotation categories specified by the prompt instructions. These ``edge-case'' annotations can then be manually reviewed by humans.

One approach is to exploit the generative LLMs' predicted token sampling process to identify an LLM's ``confidence'' in an annotation, measured by the consistency with which it would make the annotation if asked repeatedly. By introducing some degree of randomness in the LLM sampling process through the temperature hyperparameter and by repeating an annotation task on the same text sample, we generate an empirical measure of uncertainty in the annotation label that we deem a ``consistency score.''\footnote{Accessing token log probabilities, once available through the OpenAI API, may be an effective way to do the same type of selection approach.} Given a vector of annotations, $A$, with length $l$ for a given annotation task, \textit{consistency} is measured as the proportion of annotations that match the modal annotation category: \[
\frac{1}{l}\sum_{i=1}^l A_i = A_{mode}
\]

For the analyses conducted here, we annotate every text sample at least five times at a temperature of 0.7. Put otherwise, whether GPT-4 consistently labeled the same category across all iterations. As shown in Figure 4, annotations with a consistency of 1.0 show significantly higher accuracy (19.4\% increase), true positive rate (16.4\% increase), and true negative rate (21.4\% increase) compared to annotations with a consistency less than 1.0. Roughly 85\% of annotations had a consistency of 1.0. Therefore, consistency scores offer a useful way of identifying edge cases or more difficult annotations that may require additional human inspection.

\subsection{Manual Temperature Optimization}

Temperature, as discussed above, affects the degree of randomness introduced into LLMs' sampling process. One final way to improve LLM annotation performance may be to fine-tune this hyperparameter. Here, we report temperature fine-tuning experiments for seven of the 27 annotation tasks. In short, we annotated the same text samples numerous times using different temperatures and then compared performance across each of the temperature settings. As Figure \ref{fig:temp} shows, there is not a clear relationship between performance and temperature choice, which suggests that temperature does not play a significant role in annotation performance.

\begin{figure}[H]
    \centering
    \includegraphics[width = .48\textwidth]{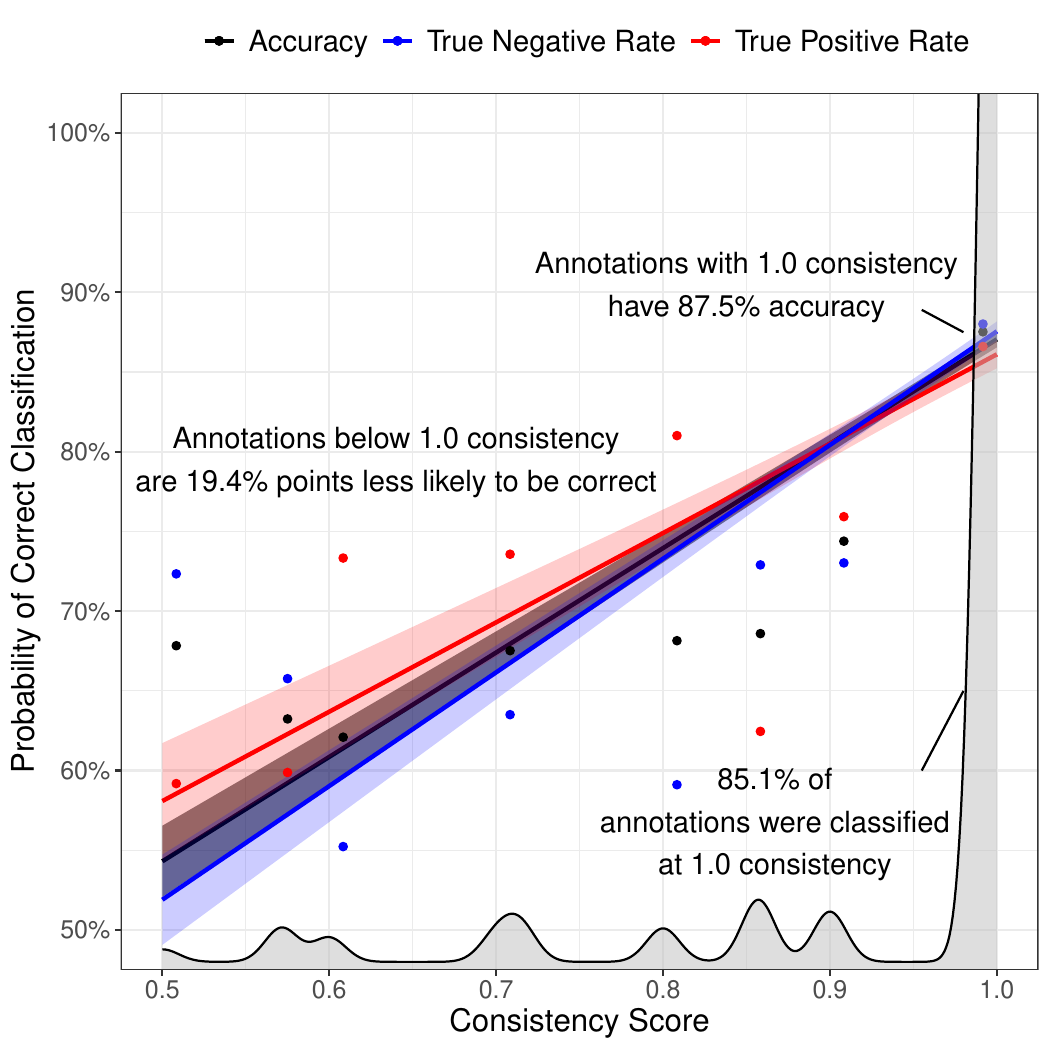}
    \caption{Relationship between consistency score and accuracy, TPR, and TNR. Lines are linear trend trends weighted by the number of samples that fall under each density score.} 
    \label{fig:codebook_updates}
\end{figure}

\subsection{GPT-4 Performance Over Time}

How does LLM performance change over time? One concerning possibility is that LLMs drift in their capabilities as they are updated over time. Some research, for example, claims that GPT-4 performance has significantly decreased since its initial release \citep{lingjiao}. This possibility is an additional reason for extra caution when deploying automated annotation tools without human validation.

To explore this possibility, we replicated a subset (14 of the 27 tasks) of our annotation tasks in both April 2023 and again in November 2023. Our results, displayed in the Appendix Figure A1, do not indicate meaningful changes in GPT-4 performance over time. If anything, Figure A1 shows a slight \textit{increase} in performance since our initial experiments. Across the 14 tasks re-analyzed, accuracy increased by 0.007 and F1 increased by 0.022 when the same annotation procedures were conducted in November 2023. Despite these minimal changes, we cannot rule out the possibility that these models will change in the future, which again underscores the importance of a human-centered framework.

\begin{figure}
    \centering
    \includegraphics[width=.9\linewidth]{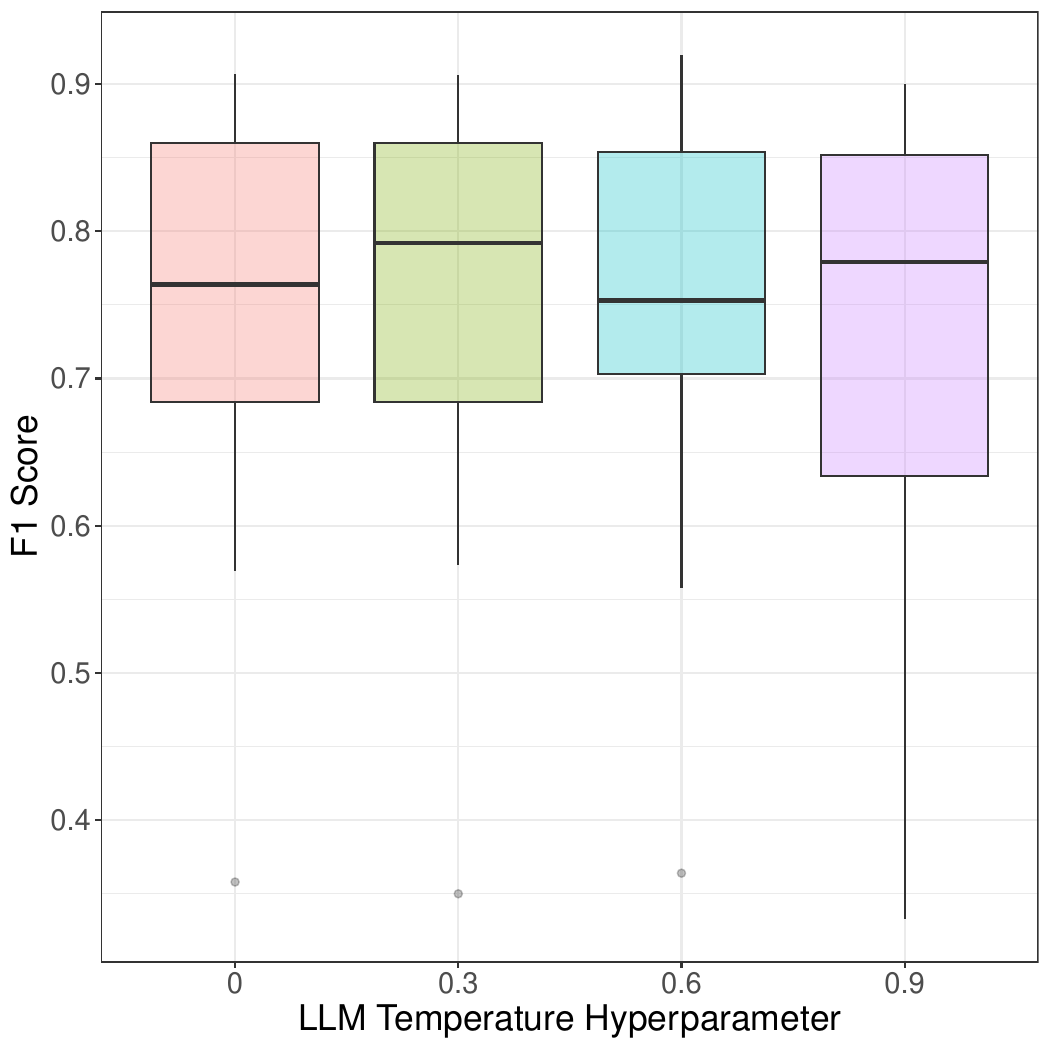}
    \caption{Relationship between temperature and F1 score.}
    \label{fig:temp}
\end{figure}

\subsection{Comparing GPT-4 to supervised model baseline}

We have shown that there are various circumstances where automated annotation with GPT-4 fails to align with human annotations. As a result, researchers should validate automated annotation quality against human-generated validation labels. Because automated annotation always requires some number of human validation text samples, it may be preferable in some cases to train a supervised classifier rather than simply using GPT-4 as a few-shot classification problem. Although supervised LLM classifiers (e.g., BERT) require data for fine-tuning, they are cheaper and may have similar performance to GPT-4. 

To understand how automated annotation with GPT-4 compares to a supervised LLM classifier baseline, we replicated all 27 annotation tasks with BERT classifiers fine-tuned with either 250 or 1000 training samples.\footnote{We are only able to replicate 14 of the 27 tasks with 1,000 training samples because the remaining nine studies had an insufficient number of human-labeled data.} Figure \ref{fig:supervised} reports the results of these analyses. As Figure \ref{fig:supervised} shows, GPT-4 outperforms BERT in 18 of the 27 tasks when there are only 250 training samples to fine-tune BERT. This is likely due to the fact that BERT is unable to learn an adequate representation of the classes with few samples, especially in the presence of class imbalance. When there are minimal training samples, GPT-4 may be preferable to a supervised classifier. Nevertheless, BERT surpasses GPT-4 performance as more training data are added to the model. Specifically, BERT performs better than GPT-4 on 10 of the 14 replicated tasks when there are 1,000 training samples used to fine-tune BERT. Researchers, especially those with cost restraints, should still consider the feasibility of using supervised classification for automated annotation.

\section{Discussion}

Given their speed, relatively low cost, and ease of application, it is tempting to deploy LLMs for automated annotation \textit{without any comparison to humans}. We caution against this approach in CSS and other research areas. If practitioners deploy LLMs for automated annotation without validating performance, they risk inaccuracies or pervasive bias. Other domains, like healthcare, have already documented concerning instances of bias in the deployment of LLMs. \citet{lancet}, for example, find that GPT-4 in applied clinical and medical applications, ``can propagate, or even amplify, harmful societal biases, raising concerns about the use of GPT-4 for clinical decision support." Other domains, like algorithmic fairness \citep{felkner-etal-2024-gpt}, emphasize similar types of concerns in automated annotation with LLMs. As a result, researchers across domains must ground generative AI in human judgment by validating against labels that have been prepared by annotators trained to anticipate and avoid potential biases. 
 
 By comparing performance across tasks and modeling strategies, we offer several advancements to the emerging literature on automated annotation using generative LLMs. First, we find that LLMs can offer high-quality labels on a wide variety of annotation tasks. Yet, automated annotation performance is subpar on a large number of tasks, which reinforces the necessity of creating a human-labeled validation set for each new annotation task. These types of procedures can help to test for convergent validity \citep{jacobs2021measurement}. Second, recall is higher than precision on the vast majority of replicated annotation tasks. Third, various techniques, such as prompt optimization, do not improve performance enough to overcome concerns of poor performance. Fourth, while GPT-4 outperforms supervised LLM classifiers when there are minimal training samples, supervised classifiers quickly surpass GPT-4’s performance as additional training data are added to the supervised model.
 \looseness=-1

In our analyses, we find significant heterogeneity in LLM performance across a range of CSS annotation tasks. Although we are hesitant to suggest that LLMs are better at certain types of tasks than others, our results indicate that simple, summarization tasks are currently better suited for automated annotation rather than nuanced tasks that require cultural context. Because of the idiosyncrasies associated with specific tasks and datasets, however, there are circumstances where LLMs fail to deliver accurate results due to ineffective prompts, noisy text data, or difficult annotation tasks. Because these challenges will persist even as LLM technology improves,\footnote{Even as LLM technology increases in sophistication, humans could still provide ambiguous prompt instructions, which could negatively affect LLM performance on an annotation task.} researchers using LLMs for annotation must always validate on a task-by-task basis. Rigorous validation is necessary to help researchers craft effective prompts and determine whether LLM annotation is viable for their annotation tasks and datasets.


Future work should investigate how to effectively evaluate automated annotation with LLMs when human validation is difficult. Human annotation is especially challenging for conceptual categories that are nuanced, contested, and subjective. Prior work has shown the challenges with human coding efforts \citep{mikhaylov2012coder}, which underscores the necessity of first validating human labels through inter-coder agreement metrics \citep{artstein2008inter}. In some cases, it may also be appropriate to model uncertainty in the human labels \citep{davani2022dealing, keith2020uncertainty}. Across these situations, researchers should adopt language from measurement modeling frameworks to guide their evaluation approaches \citep{jacobs2021measurement}. However, if humans cannot effectively generate validation data then there may be concerns whether a statistical model can reasonably complete the same task either. Foregoing validation against human labels assumes that LLM judgment on an annotation task matches human judgment. The results of this study demonstrate that such an assumption is a serious concern in practice.

\bibliography{main}

\clearpage

\section{Limitations}

We discuss several limitations of our analysis. First, our analyses primarily focus on closed-source models, although Appendix D briefly investigates an open-source alternative. While future work should continue to explore open-source automated annotations, the main thrust of our argument and analysis does not change. Either way, automated annotation should always be human-centered in some capacity. 

Second, considering human labels as ground truth is an additional possible limitation. While our annotation sources deployed experts or multiple humans, it is not impossible that these human annotators made correlated errors. Thus, some disagreements between human labels and GPT-4 annotations may come from human error. These errors would bias performance downward. This is part of the reason we compared our results to a supervised classifier as well, since both of these approaches would suffer from ground truth errors. Providing a comparison between these two approaches allows for performance comparison with any ground truth errors in the data. In Appendix E, we explore the correlation between the human label quality (i.e., intercoder agreement) and performance across tasks.

For our analyses, however, the replicated annotation tasks and data come from peer-reviewed research in high-impact journals. We believe that this helps to reduce concerns about data annotation quality. The annotation procedures in these tasks were approved by IRB protocols and were scrutinized by independent reviewers for publication in a high-impact journal. Nevertheless, it is critical to acknowledge that researchers should collect high-quality human labels and validate them as well \citep{artstein2008inter}.

\section{Ethics Statement}

In addition to adding the mandatory Paper Checklist (added after all appendices), we also include a brief ethics statement. Our research positively contributes to society and human well-being without violating any social contracts. We test generative AI tools that can help computational social scientists investigate the social world. Using the methods we explore will help researchers better understand a wide range of challenging social problems. Moreover, because our analyses require significantly fewer resources than alternatives, we believe that our results can help reduce inequities in resources across scientists.

Because of the potential negative impacts of deploying biased models, we stress the importance of human validation throughout our paper. We argue that it is critical to develop rigorous evaluation standards that are human-centered. This is part of the reason why we centered our experiments on data sets and tasks less prone to contamination. 

We also take steps to reduce potential negative impacts of this research. Specifically, we respected confidentiality and privacy issues. We also adhered to each data repository’s usage and replication policies. Moreover, all of the original studies received IRB approval and our experiments conformed to the same safety protocols. All data was anonymized by the original publication authors. Appendix B provides more details on human annotation protocols, which were conducted by the original authors and received IRB approval.

\clearpage

\appendix

\label{sec:appendix}
\setcounter{table}{0}
\setcounter{figure}{0}
\renewcommand{\thetable}{A\arabic{table}}
\renewcommand{\thefigure}{A\arabic{figure}}

\clearpage

\section{Appendix A: Replication data and tasks}

\begin{table}[H]
    \centering
    \begin{tabular}{p{3cm}p{6cm}p{4cm}p{1cm}}
         \toprule
         Author(s) & Title & Journal & Year \\
         \midrule
          Busby, and Gubler, Hawkins & Framing and blame attribution in populist rhetoric & Journal of Politics & 2019 \\
         \midrule

         Card et al. & Computational analysis of 140 years of US political speeches reveals more positive but increasingly polarized framing of immigration & PNAS & 2022 \\
         \midrule

         Cusimano and Goodwin &  People judge others to have more voluntary control over beliefs than they themselves do & Journal of Personality and Social Psychology & 2020 \\
         \midrule

         Gohdes & Repression Technology: Internet Accessibility and State Violence & American Journal of Political Science & 2020 \\
         \midrule
         Hopkins, Lelkes, and Wolken & The Rise of and Demand for Identity-Oriented Media Coverage & American Journal of Political Science & 2024 \\
         \midrule

         Müller & The Temporal Focus of Campaign Communication & Journal of Politics & 2021 \\
         \midrule
         
         Peng, Romero, and Horvat & Dynamics of cross-platform attention to retracted papers & PNAS & 2022 \\
         \midrule
         Saha  et al. & On the rise of fear speech in online social media  & PNAS & 2022 \\
         \midrule
         Schub & Informing the Leader: Bureaucracies and International Crises & American Political Science Review & 2022 \\
         \midrule
         Wojcieszak et al. & Most users do not follow political elites on Twitter; those who do show overwhelming preferences for ideological congruity & Science Advances & 2022 \\
         \midrule

         Yu and Zhang & The Impact of Social Identity Conflict on Planning Horizons & Journal of Personality and Social Psychology & 2022 \\
         \bottomrule
         & 
    \end{tabular}
    \caption{Sources of annotation tasks replicated in analysis.}
    \label{tab:articles}
\end{table}

\begin{table*}[t]
\centering
\begin{tabular}{p{2.5cm}p{8cm}p{1.5cm}}
\toprule
\multicolumn{1}{c}{Study}         & \multicolumn{1}{c}{Annotation tasks}        & \multicolumn{1}{c}{Annotator(s)}                                                                                                                                                                     \\ \midrule
Busby, Gubler, \& Hawkins (2019)  & Label open-ended text from participants in experiment for whether they (1) attribute blame to a specific actor, (2) attribute blame to a nefarious elite actor, or (3) include a positive mention of the collective people & Research assistant \\ \midrule

Card et al. (2022)                & Label congressional speeches for whether they are about immigration; also label tone as (1) proimmigration, (2) antiimmigration, or (3) neutral & Research assistant                                                        \\ \midrule

Cusimano \& Goodwin (2020)        &  Label open-ended text on climate change from participants in experiment for the presence of (1) generic reasoning about beliefs and (2) supporting evidence for the belief & Research assistant                                         \\ \midrule

Gohdes (2020)                     & Label Syrian death records based on type of killing: targeted or untargeted & Expert                                                                                                                          \\ \midrule
Hopkins, Lelkes, \& Wolken (2023) & Label if news content (headline, tweet, or Facebook share blurb) references social groups defined by (1) race/ethnicity; (2) gender/sexuality; (3) politics; (4) religion & Crowd \\ \midrule

Müller (2021)                     & Label sampled sentences from political party manifestos for if the temporal direction is (1) past, (2) present, or (3) future & Expert                                                                                                                                                                                                            \\ \midrule

Peng, Romero, \& Horvat (2022)    & Label tweets for whether they criticize academic papers  & Expert                                                                                                                 \\ \midrule
Saha et al. (2020)                & Label Gab posts for whether they include (1) fear speech and/or (2) hate speech & Crowd                           \\ \midrule

Schub (2020)                      & Label texts from Cold War crises as political or military & Expert                                                                                                                  \\ \midrule

Wojcieszak et al. (2022)          & Label a quote tweet as (1) negative, (2) neutral or (3) positive toward the message and/or the political actor being quoted, independently of the tone of the original message   & Expert                                                \\ \midrule

Yu \& Zhang (2023)                & Label  open-ended text about plans for the future from participants in experiment for whether they are about the (1) proximate future or (2) distant future   & Research assistant      \\ \bottomrule
\end{tabular}
\caption{Descriptions of annotation tasks replicated in analysis.}
\label{tab:articles2}
\end{table*}
 
\clearpage

\section{Appendix B: Human labels in replication data}

We use a novel dataset for this analysis, compiled using human-labeled data from recent studies in high-impact journals (see \ref{tab:articles}). Rather than attempting to assess the quality of the annotation procedure used by the original authors, we rely on the peer-reviewed publication process as an indication that these studies are at least of representative quality of computational social science research.

The human-labeled classification tasks in these studies are ones that can be tested without concerns of contamination. Although the raw text data (e.g., Facebook posts and tweets) are possibly included the LLM pretraining data, the associated labels from the original annotators are certainly not in the pretraining data. This is because the human-generated labels accompanying each text sample (e.g., whether a specific tweet referenced a gender identity frame) are not publicly available. Because the text \textit{with the human labels} is not included in the pretraining data, there is minimal cause for contamination concerns.

To replicate their annotation procedures, we use verbatim (or, if the original materials are not available, paraphrase) the annotation instructions used in the original study for our initial prompts. All prompt instructions are included in the supplementary material. The human coders for these datasets were experts, undergraduate RAs, or crowd-sourced workers. The original authors had received IRB approval for each study we replicate. Further details about the annotation procedures for each replicated study can be found in the methodological details for those studies.

\section{Appendix C: Methodological details}

To establish a baseline for classification performance, we fine-tuned BERT classifiers with varying training data strategies: n=250 human samples and n=1,000 human samples. We use BERT as it is among the most popular models used for supervised classification in computational social science. Because the hyperparameters that yield strongest performance may differ across tasks, we used a grid search. For each classifier, we searched over 18 combinations of learning rate (1e-5, 2e-5, and 5e-5), batch size (8 and 16), and number of epochs (2, 4, 6). This grid search was conducted on a subsample of 250 text samples per task. We identified the combination of hyperparameters with the highest F1 and retained that combination for all models for that particular task (see best-performing hyperparameter configurations in Table \ref{tab:hyper}). Hyperparameters that remained constant across tasks and details about model architecture are shown in Table \ref{tab:arch}. We evaluate each model on 1,000 held-out text samples with labels from human annotators.

This analysis was conducted in Python 3.10.12 using HuggingFace's Transformers \citep{wolf} library and PyTorch libraries \citep{Paszke}. For data preprocessing, we use Python Pandas \citep{pandas}. We used Google Colab as our computing environment, totalling 215 T4 GPU compute units (approximately 421.4 GPU hours). In the supplementary material, we include all code to run our supervised training procedures. 

\section{Appendix D: Comparing to an open-source alternative}

Due to various concerns about closed-source models (e.g., Spirling, 2023), we also generate few-shot labels using Mistral-7B from the same procedures employed in the GPT-4 few-shot model. We include the Mistral-7B few-shot model as a smaller, open-source alternative to GPT-4. Mistral-7B was selected because the model weights are available for download and it has higher performance than Llama-13B \citep{jiang2023mistral}.

Figure \ref{fig:open} compares automated annotation classification performance between GPT-4 and Mistral-7B. The evaluation sets for these experiments are the same as the main analyses previously conducted. Figure \ref{fig:open} shows that there is a fairly sizeable gap between the open-source (Mistral-7B) and closed-source (GPT-4) few-shot models. F1 scores for Mistral-7B are 0.16 worse, on average, than GPT-4, although this may be expected from a significantly smaller and free-to-use model. Cheaper, smaller open models may perform worse on automated annotation than their larger, propriety counterparts. Thus, these findings further reinforce the necessity of human validation to confirm annotation quality.

\section{Appendix E: Relationship between task difficulty and performance}

Here, we investigate automated annotation performance across different text and task difficulties. Figure \ref{fig:difficulty} measures text and task complexity in different ways and compares difficulty to GPT-4 annotation performance. The top two panes in Figure \ref{fig:difficulty} show that text length and text readability (Flesch-Kincaid grade level score)\footnote{The Flesch-Kincaid Grade Level metric signals how challenging a text passage is for a human to read. The numeric score translates to the recommended grade level required to read the text. The score is calculated based on the average number of words per sentence and average number of syllables per word.} are uncorrelated to automated annotation performance. Both of these measures were calculated on the text samples in the original studies. Neither of these measures of text difficulty predict poor performance.

The bottom pane of Figure \ref{fig:difficulty} shows the correlation between intercoder reliability and GPT-4 F1 performance. Across the 27 replicated tasks, 21 tasks in the original studies reported metrics on intercoder reliability. 16 studies utilized Krippendorff’s alpha, while five studies calculated Cohen's kappa. If the original study involved a multi-class task, a single reliability metric was noted in the study, pooled across each label. Figure \ref{fig:difficulty} illustrates that, when intercoder reliability is low, GPT-4 also has low performance. This could possibly be due to inadequate conceptualization of the annotation category, which might lead to inconsistent guidelines for sample categorization. Categories with medium to high intercoder reliability display higher performance than the categories with low intercoder reliability. Beyond this, however, there is minimal correlation between high intercoder reliability and strong GPT performance. Thus, while it is possible that poor performance was the result of difficult coding tasks, the evidence provided here suggests that strong intercoder reliability only weakly predicts strong performance.

\section{Appendix F: Miscellaneous additional information}

Additional sources:
\begin{itemize}
    \item Robot image (used in Figure 1): \url{https://commons.wikimedia.org/wiki/File:Grey_cartoon_robot.png}
    \item Human silhouette image (used in Figure 1): \url{https://commons.wikimedia.org/wiki/File:SVG_Human_Silhouette.svg}
\end{itemize}

\begin{figure*}
    \centering
    \includegraphics[width = .8\textwidth]{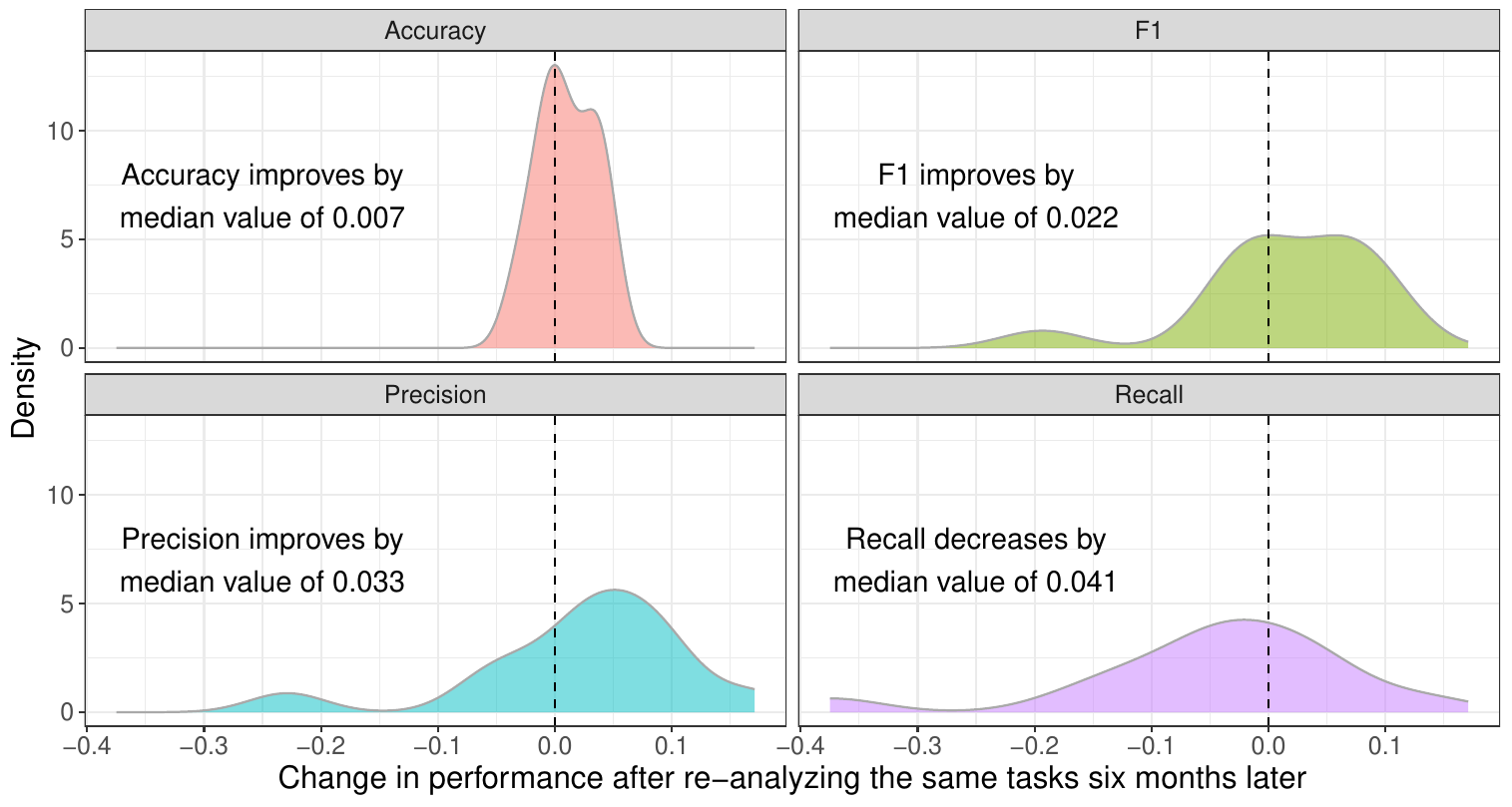}
    \caption{Examining GPT-4 performance over time}
    \label{fig:A7}
\end{figure*}

\begin{figure*}
    \centering
    \includegraphics[width = .85\textwidth]{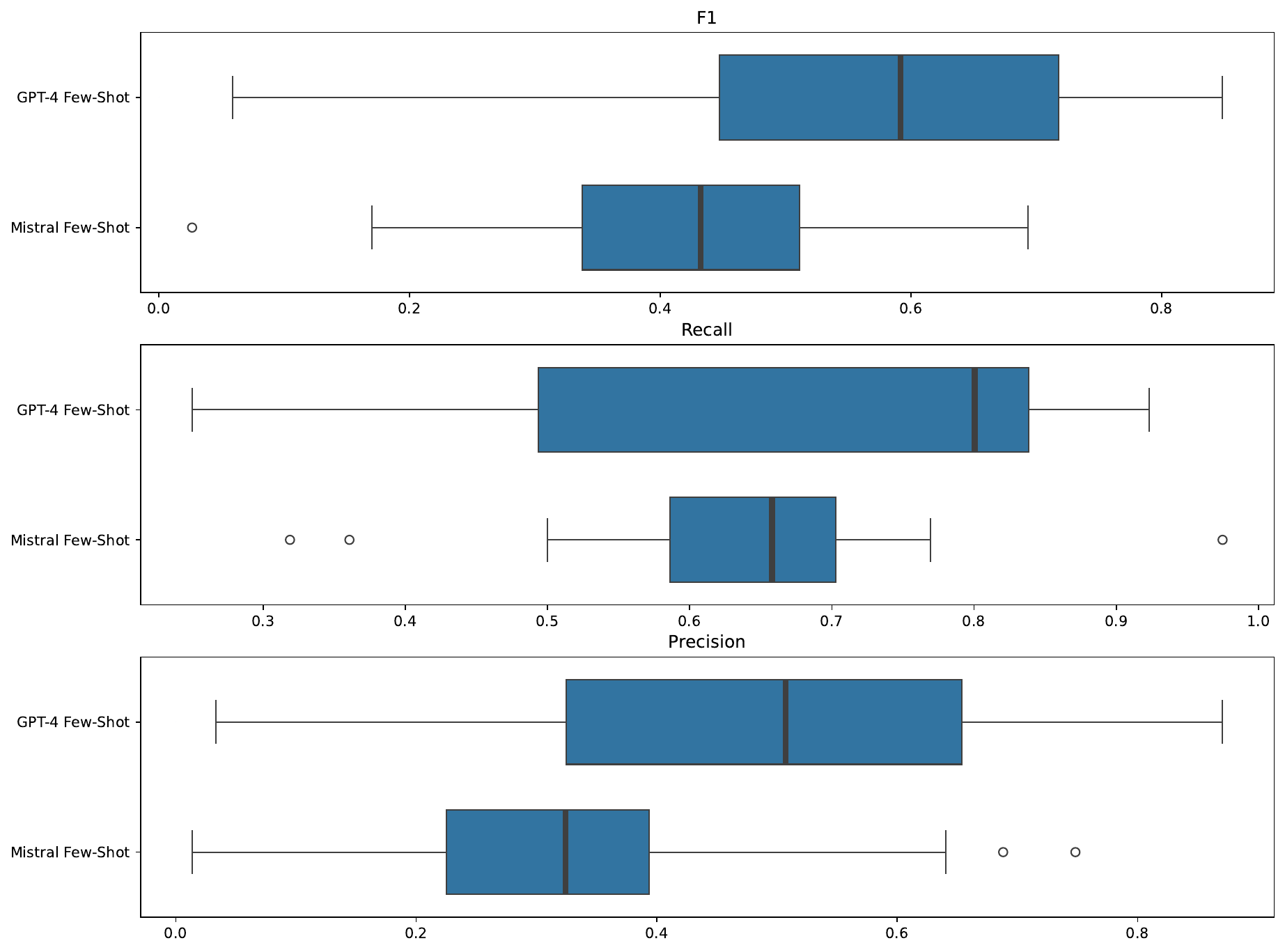}
    \caption{Comparing GPT-4 and Mistral-7B few-shot classifiers}
    \label{fig:open}
\end{figure*}

\begin{figure*}
    \centering
    \includegraphics[width = .8\textwidth]{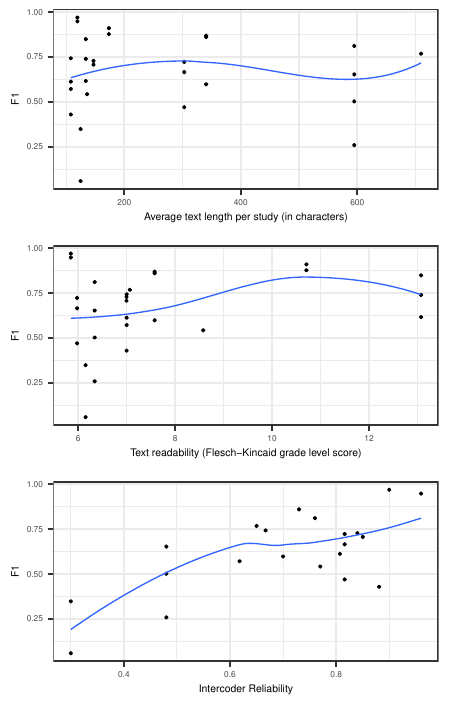}
    \caption{Relationship between text/task difficulty and performance.}
    \label{fig:difficulty}
\end{figure*}

\begin{figure*}
    \centering
    \includegraphics[width = 1\textwidth]{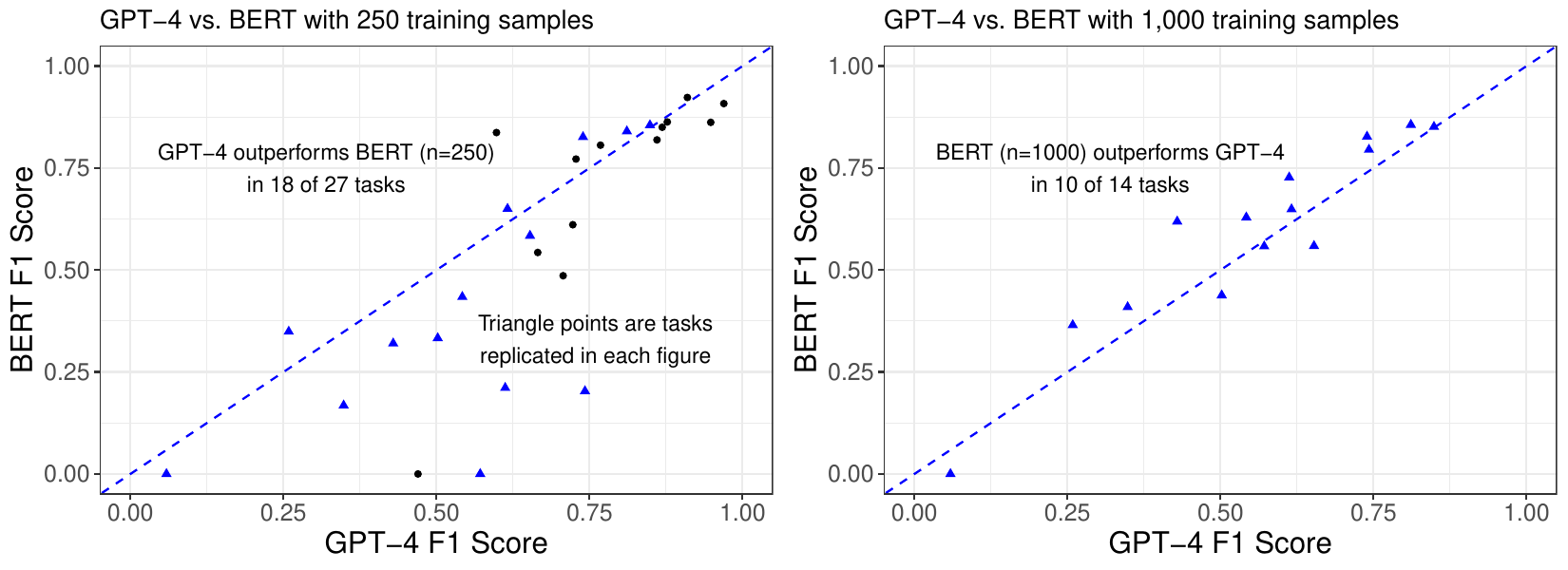}
    \caption{Supervised text classifiers outperform GPT-4 with additional training samples.}
    \label{fig:supervised}
\end{figure*}

\begin{table*}[b] 
    \centering
    \begin{tabular}{p{2.2 cm}|p{5.5cm}p{6.7cm}}
         \toprule
         Study & Task & Hyperparameters \\
         \midrule
         \multirow{3}{*}{Busby} & Classify specific blame & learning rate (5e-05), batch size (8), epochs (6) \\
         \cmidrule(lr){2-3}
              & Classify elite blame  & learning rate (5e-05), batch size (8), epochs (4) \\
         \cmidrule(lr){2-3}
              & Classify collective positive & learning rate (5e-05), batch size (8), epochs (6) \\
   
         \midrule
              \multirow{2}{*}{Cusimano et al.} & Classify generic belief & learning rate (5e-05), batch size (8), epochs (6) \\
         \cmidrule(lr){2-3}
              & Classify evidence belief & learning rate (2e-05), batch size (16), epochs (2) \\
        \midrule
            \multirow{4}{*}{Card et al.} & Classify immigration speeches & learning rate (5e-05), batch size (8), epochs (4) \\
            \cmidrule(lr){2-3} 
              & Classify pro-immigration speeches & learning rate (5e-05), batch size (16), epochs (6) \\
              \cmidrule(lr){2-3}
            & Classify anti-immigration speeches & learning rate (5e-05), batch size (8), epochs (6) \\
            \cmidrule(lr){2-3}
             & Classify neutral immigration speeches & learning rate (5e-05), batch size (8), epochs (4) \\
        \midrule
        \multirow{2}{*}{Gohdes} & Classify targeted killings & learning rate (5e-05), batch size (8), epochs (4) \\
         \cmidrule(lr){2-3}
              & Classify untargeted killings & learning rate (5e-05), batch size (16), epochs (6) \\
        
         \midrule
            \multirow{4}{*}{Hopkins et al.} & Classify race/ethnicity & learning rate (2e-05), batch size (8), epochs (4) \\
            \cmidrule(lr){2-3}
              & Classify gender & learning rate (5e-05), batch size (8), epochs (6) \\
         \cmidrule(lr){2-3}
              & Classify political groups & learning rate (5e-05), batch size (16), epochs (6) \\
         \cmidrule(lr){2-3}
             & Classify religious groups & learning rate (5e-05), batch size (8), epochs (6) \\
             \midrule
         \multirow{3}{*}{Müller} & Classify past & learning rate (5e-05), batch size (8), epochs (4) \\
         \cmidrule(lr){2-3}
              & Classify present & learning rate (5e-05), batch size (8), epochs (4) \\
         \cmidrule(lr){2-3}
              & Classify future & learning rate (2e-05), batch size (8), epochs (6) \\
         \midrule
         Peng et al. & Classify criticism & learning rate (5e-05), batch size (8), epochs (6) \\
         \midrule
       \multirow{2}{*}{Saha  et al.} & Classify fear speech & learning rate (5e-05), batch size (8), epochs (6) \\
         \cmidrule(lr){2-3}
              & Classify hate speech & learning rate (5e-05), batch size (8), epochs (4) \\
        \midrule
         Schub & Classify political or military text & learning rate (5e-05), batch size (8), epochs (6) \\
         \midrule
          \multirow{3}{*}{Wojcieszak et al.} & Classify positive & learning rate (2e-05), batch size (8), epochs (6) \\
         \cmidrule(lr){2-3}
              & Classify negative & learning rate (5e-05), batch size (16), epochs (6) \\
         \cmidrule(lr){2-3}
              & Classify neutral & learning rate (2e-05), batch size (8), epochs (2) \\
         \midrule
              \multirow{2}{*}{Yu et al.} & Classify proximate future & learning rate (5e-05), batch size (8), epochs (6) \\
         \cmidrule(lr){2-3}
              & Classify distant future & learning rate (5e-05), batch size (8), epochs (6) \\
         \bottomrule
    \end{tabular}
    \caption{Hyperparameter settings per task.}
    \label{tab:hyper}
\end{table*}

\begin{table*}[b] 
    \centering
    \begin{tabular}{p{4 cm}p{2 cm}}
         \toprule
          & BERT-base \\
         \midrule
            \# parameters & 110m \\
         \midrule
         \# attention heads & 12 \\
         \midrule
          Hidden dim. & 768 \\
         \midrule
         Feedforward dim. & 3072 \\
         \midrule
         Activation  & GELU \\
         \midrule
         Dropout  & 0.1 1\\
         \midrule
         Optimizer  & Adam \\
         \midrule
         Weight decay  & 0.01 \\
         \bottomrule
         & 
    \end{tabular}
    \caption{Model architectures and additional hyperparameters.}
    \label{tab:arch}
\end{table*}

\clearpage

\section{Checklist}

\begin{enumerate}

\item For most authors...
\begin{enumerate}
    \item  Would answering this research question advance science without violating social contracts, such as violating privacy norms, perpetuating unfair profiling, exacerbating the socio-economic divide, or implying disrespect to societies or cultures?
    \answerYes{Yes, see Ethics Statement}
  \item Do your main claims in the abstract and introduction accurately reflect the paper's contributions and scope?
    \answerYes{Yes, see Results}
   \item Do you clarify how the proposed methodological approach is appropriate for the claims made? 
    \answerYes{Yes, see Replication Procedures}
   \item Do you clarify what are possible artifacts in the data used, given population-specific distributions?
    NA
  \item Did you describe the limitations of your work?
    \answerYes{Yes, see Limitations}
  \item Did you discuss any potential negative societal impacts of your work?
    \answerYes{Yes, see Ethics Statement}
      \item Did you discuss any potential misuse of your work?
    \answerYes{Yes, see Ethics Statement}
    \item Did you describe steps taken to prevent or mitigate potential negative outcomes of the research, such as data and model documentation, data anonymization, responsible release, access control, and the reproducibility of findings?
    \answerYes{Yes, see Ethics Statement}
  \item Have you read the ethics review guidelines and ensured that your paper conforms to them?
    \answerYes{Yes}
\end{enumerate}

\item Additionally, if your study involves hypotheses testing...
\begin{enumerate}
  \item Did you clearly state the assumptions underlying all theoretical results?
    NA
  \item Have you provided justifications for all theoretical results?
    NA
  \item Did you discuss competing hypotheses or theories that might challenge or complement your theoretical results?
    NA
  \item Have you considered alternative mechanisms or explanations that might account for the same outcomes observed in your study?
    NA
  \item Did you address potential biases or limitations in your theoretical framework?
    NA
  \item Have you related your theoretical results to the existing literature in social science?
    NA
  \item Did you discuss the implications of your theoretical results for policy, practice, or further research in the social science domain?
    NA
\end{enumerate}

\item Additionally, if you are including theoretical proofs...
\begin{enumerate}
  \item Did you state the full set of assumptions of all theoretical results?
    NA
	\item Did you include complete proofs of all theoretical results?
    NA
\end{enumerate}

\item Additionally, if you ran machine learning experiments...
\begin{enumerate}
  \item Did you include the code, data, and instructions needed to reproduce the main experimental results (either in the supplemental material or as a URL)?
    \answerYes{Yes, attached as supplementary materials}
  \item Did you specify all the training details (e.g., data splits, hyperparameters, how they were chosen)?
    \answerYes{Yes, see Appendix C}
     \item Did you report error bars (e.g., with respect to the random seed after running experiments multiple times)?
    NA
	\item Did you include the total amount of compute and the type of resources used (e.g., type of GPUs, internal cluster, or cloud provider)?
    \answerYes{Yes, see Appendix C}
     \item Do you justify how the proposed evaluation is sufficient and appropriate to the claims made? 
    \answerYes{Yes, see Introduction, Results}
     \item Do you discuss what is ``the cost`` of misclassification and fault (in)tolerance?
    \answerYes{Yes, see Introduction, Discussion}
  
\end{enumerate}

\item Additionally, if you are using existing assets (e.g., code, data, models) or curating/releasing new assets, \textbf{without compromising anonymity}...
\begin{enumerate}
  \item If your work uses existing assets, did you cite the creators?
    \answerYes{Yes}
  \item Did you mention the license of the assets?
    NA
  \item Did you include any new assets in the supplemental material or as a URL?
    NA
  \item Did you discuss whether and how consent was obtained from people whose data you're using/curating?
    \answerYes{Yes, see Appendix B}
  \item Did you discuss whether the data you are using/curating contains personally identifiable information or offensive content?
    \answerYes{Yes, see Ethics Statement and Appendix B}
\item If you are curating or releasing new datasets, did you discuss how you intend to make your datasets FAIR?
NA
\item If you are curating or releasing new datasets, did you create a Datasheet for the Dataset (see \citet{gebru2021datasheets})? 
NA
\end{enumerate}

\item Additionally, if you used crowdsourcing or conducted research with human subjects, \textbf{without compromising anonymity}...
\begin{enumerate}
  \item Did you include the full text of instructions given to participants and screenshots?
    \answerNo{No, because we do not work directly with human subjects from the original study. In the supplementary materials, we include LLM prompt instructions that are based on instructions given to humans.}
  \item Did you describe any potential participant risks, with mentions of Institutional Review Board (IRB) approvals?
    \answerNo{No, because we do not work directly with human subjects from the original study. All replicated studies received IRB approval.}
  \item Did you include the estimated hourly wage paid to participants and the total amount spent on participant compensation?
    \answerNo{No, because we do not work directly with human subjects from the original study.}
   \item Did you discuss how data is stored, shared, and deidentified?
   \answerNo{No, because we are not publicly releasing the data without the original authors permission. All data was already deidentified.}
\end{enumerate}

\end{enumerate}

\end{document}